\documentclass{article}

\usepackage[dblblindworkshop,final]{neurips_2026}
\workshoptitle{Continual Learning for Enterprise AI Agents (CLEA)}

\usepackage[utf8]{inputenc}
\usepackage[T1]{fontenc}
\usepackage{hyperref}
\usepackage{url}
\usepackage{booktabs}
\usepackage{amsfonts}
\usepackage{amsmath}
\usepackage{nicefrac}
\usepackage{microtype}
\usepackage{xcolor}
\usepackage[normalem]{ulem}
\usepackage{graphicx}
\usepackage{multirow}
\usepackage{algorithm}
\usepackage{float}
\usepackage{algpseudocode}
\usepackage{enumitem}

\renewcommand{\topfraction}{0.92}
\renewcommand{\textfraction}{0.06}
\renewcommand{\floatpagefraction}{0.75}
\newcommand{\sysname}{REM}
\newif\iftrackchanges
\IfFileExists{trackmode.tex}{\trackchangesfalse
}{\trackchangesfalse}

\iftrackchanges
  \usepackage[xetex]{changebar}
  \cbcolor{red}
  \DeclareRobustCommand{\rev}[1]{\cbstart{\color{red}\relax#1}\cbend\relax}
  \DeclareRobustCommand{\del}[1]{\cbstart{\color{red}\relax\sout{#1}}\cbend\relax}
  
  \DeclareRobustCommand{\todo}[1]{\cbstart\fcolorbox{red}{yellow}{\parbox[t]{0.94\linewidth}{\scriptsize\textbf{TODO:} #1}}\cbend\relax}
  \newcommand{\trackbanner}{%
    \begin{center}
    \fcolorbox{red}{red!8}{\parbox{0.93\linewidth}{\footnotesize\centering
      \textbf{TRACKED-CHANGES COPY --- not for upload.}
      Edits since the submitted version are \textcolor{red}{in red with a change bar in the margin};
      deletions are \textcolor{red}{\sout{struck through}}; yellow boxes are open questions for the authors.
      The clean PDF in the same folder is the upload candidate.}}
    \end{center}\medskip}
\else
  \DeclareRobustCommand{\rev}[1]{#1}
  \DeclareRobustCommand{\del}[1]{\ignorespaces}
  
  \DeclareRobustCommand{\todo}[1]{}
  \newcommand{\trackbanner}{}
\fi

\title{\sysname{}: What to Store, When to Recall, What to Forget\\in Experience Memory for Coding Agents}

\author{%
  \textbf{Xinyi Lisa Chen}$^{1*\ddagger}$ \quad \textbf{Pedro Herrero-Vidal}$^{2*\dagger}$ \quad \textbf{Xiaohan Chen}$^{2}$ \quad \textbf{Xiaojing Yu}$^{2}$ \\
  \textbf{Cris Liu}$^{2}$ \quad \textbf{Bingqing Song}$^{2}$ \quad \textbf{Bin Xu}$^{2}$ \quad \textbf{Suleiman A. Khan}$^{2}$ \quad \textbf{You-Lin Chen}$^{2\dagger}$ \\[0.6em]
  $^{1}$Yale University \qquad $^{2}$Amazon \\
  \texttt{xinyi.lisa.chen@yale.edu} \\
  \rev{\texttt{\{phvidal, chxh, yyxjing, licris\}@amazon.com}} \\
  \rev{\texttt{\{bingqiso, binxu, suleimkh, cyoulin\}@amazon.com}}
}

\begin{document}

\maketitle
\def\thefootnote{*}\footnotetext{These authors contributed equally to this work.}
\def\thefootnote{$\dagger$}\footnotetext{Corresponding authors.}
\def\thefootnote{$\ddagger$}\footnotetext{Work done while an intern at Amazon.}
\def\thefootnote{\arabic{footnote}}
\setcounter{footnote}{0}

\trackbanner

\begin{abstract}
Large language model (LLM) agents now autonomously complete long-horizon tasks, such as repository-level coding, through hundreds of think-act steps, yet the experience each run produces is rarely reused on the next. Prior memory systems concentrate on retrieval, finding the most relevant past trajectory. Retrieval cannot observe the running agent, so it cannot decide whether the agent needs an experience, at which point, or for how long. We introduce \sysname{} (Reusable Experience Memory), a framework that manages the memory lifecycle from inside the agent's think-act loop through three explicit decisions. \emph{What to store:} each completed trajectory is consolidated into a phase-structured document, two orders of magnitude smaller than the trajectory. \emph{When to recall:} the document is injected at the start of the task; the costlier implementation record is withheld until an in-loop gate detects the agent struggling. \emph{What to forget:} the record is evicted after a single exposure. On repository-level bug fixing (SWE-ContextBench Lite), \sysname{} retrieves the related trajectory more accurately than every state-of-the-art memory system (68.7\% vs.\ 55.6\% match@1) and \del{consistently improves}\rev{improves, in every replicate,} end-task resolution (27.6\% vs.\ 24.9\%) at 17\% lower cost. Injecting the same experience \del{without a policy}\rev{as a summary or raw trajectory}, even when oracle-retrieved, does not. Ablating any of the three decisions reduces resolution. Altogether, these results demonstrate that the bottleneck in experience reuse is not retrieval fidelity but \rev{how experience is stored and governed across its} lifecycle\del{ governance}, opening a control-theoretic perspective on agent memory.
\end{abstract}

\section{Introduction}
\label{sec:intro}

LLM agents now run unattended for hours, solving exploratory tasks such as repository-level coding \citep{jimenez2024swebench} and text-to-SQL \citep{spider2018} through hundreds of think-act steps. Each run produces a detailed trace of exploration, dead ends, and an eventual solution, and a growing ecosystem of memory layers \citep{chhikara2025mem0, zhang2026swecontextbench} and experience-reuse frameworks \citep{shinn2023reflexion, zhao2024expel, wang2024awm, wang2023voyager} promises to make that experience compound. In practice it rarely does.

Most work targets the middle of the lifecycle: retrieving the most relevant past trajectory. Three findings from this literature suggest retrieval is not where reuse fails. First, agents use only a fraction of the context they receive, and retrieval alone is insufficient because they must actively interpret what they are given \citep{zhang2026swecontextbench}. Second, irrelevant or poorly placed context degrades performance, as models are distracted by irrelevant material and use long contexts unevenly by position \citep{shi2023distracted, liu2023lost}. Third, abstraction transfers better than detail: \citet{hu2026continual} report negative transfer from detailed trajectories, concentrated on the hard cases where guidance is most needed.

These findings share a cause: retrieval cannot observe the state of the running agent, so what happens to a retrieved experience once it enters a long-running loop is left unmanaged. This is most consequential where an agent runs repeatedly against an evolving codebase, its accumulated experience a standing asset whose compounding is the long-run aim of continual learning for agents \citep{hu2026continual}. We take one step toward that aim, reusing a fixed pool of past runs within a single new run, and find the state of the receiving agent decisive: naively recalled memories leave resolution unchanged even when an oracle supplies the correct experience, and the same memory that \del{rescues a struggling agent derails one already on track}\rev{helps mid-capability agents costs the strongest ones (\S\ref{sec:results-who})}, while agents reference only about a quarter of the identifiers in a raw trajectory and half of those in a prose summary (\S\ref{sec:results-utilization}).

We introduce \sysname{} (Reusable Experience Memory), which manages the memory lifecycle from inside the agent's think-act loop, where the agent's state is observable (Figure~\ref{fig:pipeline}), through three explicit decisions. \emph{What to store:} each trajectory is consolidated into one \emph{phase-structured document} organized by the four phases shared by diagnostic work, \emph{localize}, \emph{understand}, \emph{fix}, and \emph{verify}. It is two orders of magnitude smaller than the trajectory and serves as both the retrieval unit and the injected context, so stored tokens are spent once. \emph{When to recall:} the document is injected at task start, while the trajectory's \emph{implementation record}, far costlier and more distracting, is withheld until an in-loop node observes the agent struggling. \emph{What to forget:} the record is evicted after a single exposure\rev{, while the document stays in context for the whole run}. Every stage is a swappable policy slot, and the extraction schema, the only domain-specific artifact, is discovered automatically from example trajectories (\S\ref{sec:method-discovery}).

We evaluate on SWE-ContextBench Lite \citep{zhang2026swecontextbench}, judging memory by end-task resolution rather than retrieval accuracy alone. This work presents the following main contributions.
\begin{itemize}[leftmargin=1.2em, itemsep=2pt, topsep=3pt, parsep=0pt]
\item \textbf{From \emph{which} to \emph{when}.} The value of a recalled experience depends on the state of the receiving agent, which makes agent memory a control problem, not a lookup problem.
\item \textbf{\sysname{}.} A framework that governs the full memory lifecycle from inside the agent's loop---phased consolidation, a two-tier read path, struggle-gated recall, and one-shot eviction---with an automatically discovered extraction schema, its only domain-specific artifact (\S\ref{sec:method}).
\item \textbf{State-of-the-art retrieval; improved resolution at lower cost.} The phased document is the most accurate retrieval unit on SWE-ContextBench Lite by a 13-point margin, and \sysname{} improves resolution in every replicate while cutting cost 17\%, whereas \del{policy-free injection improves neither}\rev{start-of-run summaries and raw trajectories do not improve resolution}, even with an oracle (\S\ref{sec:results-store}, \S\ref{sec:results-main}).
\item \textbf{Decision-level ablations.} Isolating each lifecycle decision shows the stored form drives the gains, while gating and eviction keep a costly second tier from doing harm, and a utilization analysis links the effect to how much injected context agents actually use (\S\ref{sec:results-when}, \S\ref{sec:results-utilization}).
\end{itemize}

\section{Related work}
\label{sec:related}

\paragraph{Agent memory and experience reuse.} Self-improvement frameworks store verbal reflections, distilled insights, workflows, or skill libraries and read them back on later tasks \citep{shinn2023reflexion, zhao2024expel, wang2024awm, wang2023voyager}, while production memory layers shred histories into factual items or chunks behind a search interface \citep{chhikara2025mem0}. How well experience transfers depends on how it is stored, with abstract procedural memories transferring more reliably than detailed trajectories \citep{hu2026continual}. These stored units sit at the extremes of granularity, either isolated facts that lose connective structure or whole trajectories that bury the identifying tokens, and their recall is eager, delivered at the start of the next task. \rev{Structured representations of experience have also been proposed, as distilled reasoning strategies \citep{ouyang2026reasoningbank}, context-aware guidelines selected for the agent's current state \citep{fu2024autoguide}, and subtask-level memories aligned with software-engineering workflows \citep{shen2026subtask}.} \sysname{} instead consolidates each trajectory into one phase-structured document that is both the retrieval unit and the injected context\rev{, discovers its extraction schema automatically}, and evaluates it on the resolution and cost of the receiving run\rev{ against the memory systems benchmarked on the same tasks}.

\paragraph{Deciding when to retrieve and what to keep.} Adaptive retrieval learns whether a query needs retrieval at all \citep{asai2024selfrag}, and context-curation policies learn to insert and delete material from an agent's working context under a budget \citep{zhang2026memact}. Concurrent work casts memory operations as a learned control policy over when to retrieve, inject, and forget \citep{jiang2026memcon}. The machinery for deciding \emph{whether} to retrieve is well developed; what happens to retrieved experience after it enters a long-running loop is not. \sysname{} addresses that second half\rev{, the handling of experience after injection,} with the simplest sufficient mechanism, an in-loop gate that withholds costly detail until the run shows struggle and evicts it after one exposure, and reports what each decision is worth\rev{ in end-task resolution and cost}.

\paragraph{Context sensitivity and evaluation.} Language models are degraded by irrelevant input and use long contexts unevenly by position \citep{shi2023distracted, liu2023lost}, so injected experience is not free. Agentic-coding benchmarks have moved from patch correctness \citep{jimenez2024swebench} toward measuring reuse, and SWE-ContextBench reports that correctly selected summaries help while unfiltered context does not, concluding that agents must interpret what they receive \citep{zhang2026swecontextbench}. We take that as the starting point and quantify uptake directly, measuring which injected identifiers a run references and whether that predicts resolution.

\section{An experience memory for coding agents}
\label{sec:method}

\begin{figure}[t]
\centering
\includegraphics[width=1.0\textwidth]{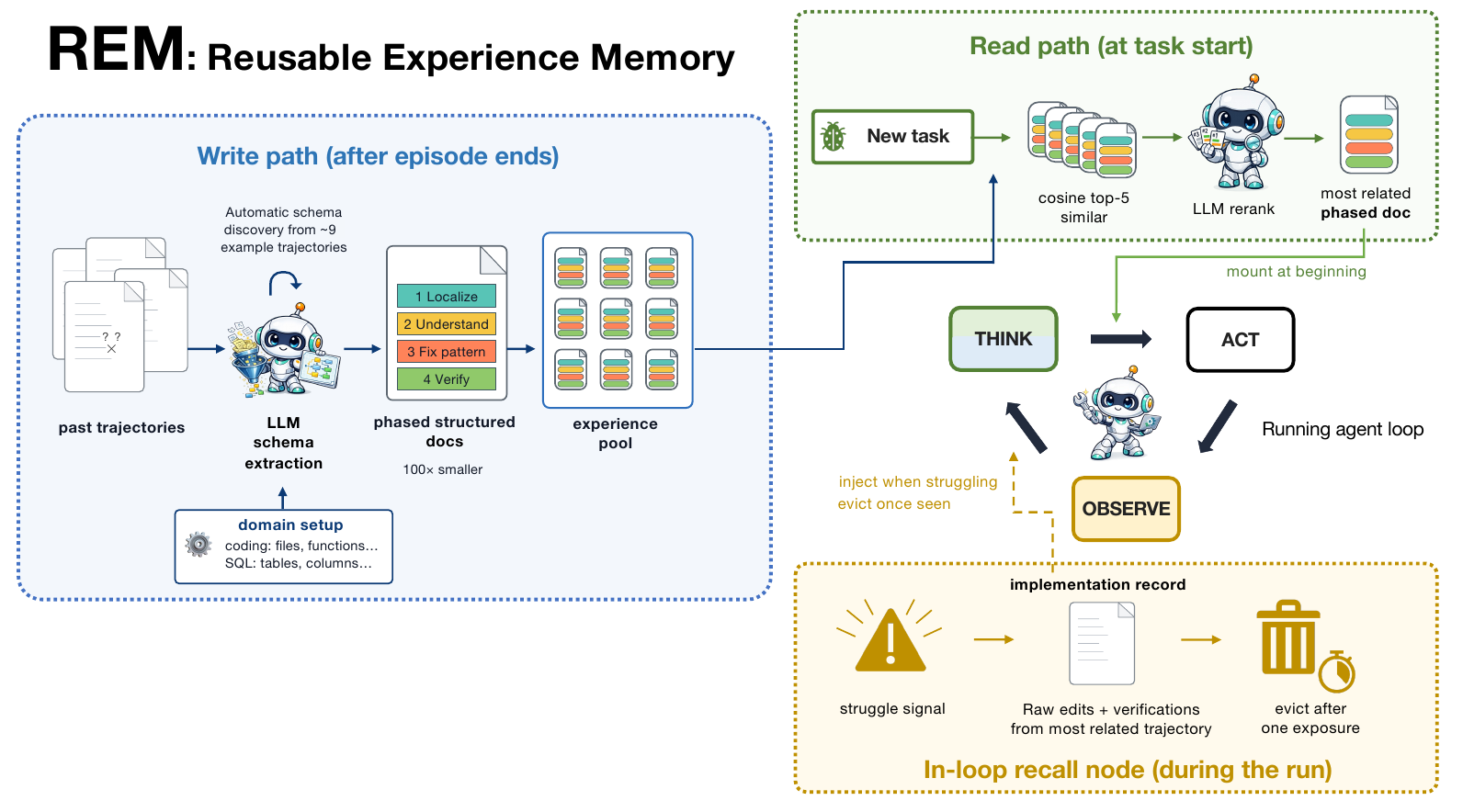}
\caption{\textbf{\sysname{} manages the memory lifecycle in three stages around the agent's think-act loop.} Left: the write path consolidates each completed trajectory into a phase-structured document (\textit{localize}, \textit{understand}, \textit{fix pattern}, \textit{verify}); the extraction fields come from a domain module discovered automatically from example trajectories. Top right: the read path retrieves candidates by cosine similarity, reranks with one LLM call, and injects the top document at the start of the run. Bottom right: the in-loop recall node recalls the implementation record on a struggle signal and evicts it after one exposure.}
\label{fig:pipeline}
\end{figure}

\subsection{Why a store needs a lifecycle policy}
\label{sec:method-why}

Reusing a trajectory is not mainly about finding it. One repository-level run produces tens of thousands of tokens, most recording abandoned paths. Stored raw, a trajectory fails twice: as a retrieval unit its few identifying tokens are diluted by exploration noise, and as injected context its size displaces the new agent's own reasoning. Shattering it into factoids removes the noise but also the structure that makes a trajectory a solution. The first decision, \textit{what to store}, is therefore the choice of a form that keeps the structure of a solution while discarding the search that produced it.

The other two decisions cannot be made by the store. Whether injected context helps depends on the state of the receiving agent: one that has already localized the defect gains little and can be derailed, whereas one that has exhausted its hypotheses stands to gain. That state is invisible at retrieval time and changes as the run proceeds, so injection must be decided while the agent works, by a component inside the loop, the second decision, \textit{when to recall}. Once injected, context left in place biases every later turn toward another run's solution, so the policy must also say when to remove it, the third decision, \textit{what to forget}.

\subsection{Framework overview}
\label{sec:method-overview}

\sysname{} wraps the loop with one stage per decision (Figure~\ref{fig:pipeline}, Algorithm~\ref{alg:rem}\rev{ in Appendix~\ref{app:algorithm}}): the write path consolidates finished trajectories into phased documents, the read path retrieves the document most related to a new task, and the in-loop node decides when the implementation record is injected and removed. The pipeline is designed for any domain whose work follows \emph{localize}, \emph{understand}, \emph{fix}, and \emph{verify}, such as debugging code, text-to-SQL repair, or incident response; the domain enters only through a swappable module of extraction fields (\S\ref{sec:method-discovery}).

\paragraph{Write path.} Each stored trajectory is a transcript of 55K--170K tokens. We flatten it to text, keeping user text, assistant text, and tool-call inputs while dropping tool results, which shrinks it roughly fifteenfold with no truncation. One LLM call extracts a schema of typed fields (for coding: files, functions, root cause, fix approach, tests), and a deterministic renderer lays them out as a four-phase document averaging 270 tokens, \del{roughly a hundredfold compression}\rev{more than two orders of magnitude smaller than the transcript}. In memory terms this is consolidation from an episodic trace into a semantic one: the solution's structure survives while the exploration noise does not.
\label{sec:method-write}

\paragraph{Read path.} We embed the pool and the new task's problem statement with an off-the-shelf text-embedding model, pass the five nearest documents by cosine similarity to one LLM rerank call, and inject the selected document. The reranker never abstains; an unparseable reply falls back to the nearest document by cosine similarity. The stage is deliberately simple: our hypothesis is that structuring what is stored matters more than how it is searched.
\label{sec:method-read}

\paragraph{Recall node: two tiers, a struggle gate, eviction.} The 270-token document is injected at the start of the run to orient the agent. The \emph{implementation record}, the raw edits and verifications from the same trajectory capped at 12K characters, is injected only on demand, the first time the agent \emph{fails a validation}, our simplest observable evidence that its own approach is not working (edits to test files are filtered so self-written tests do not count as struggle). After one exposure the record is evicted and does not persist in the cached prefix\rev{; the phased document, by contrast, stays in context for the whole run}. \S\ref{sec:results-when} ablates both halves, comparing eager against gated injection and eviction against persistence.
\label{sec:method-recall}

\subsection{Automated schema discovery and measurement}
\label{sec:method-discovery}

The extraction schema is the memory's only domain-specific artifact, and a meta-loop learns it from a few example trajectories\rev{ (nine trajectories from the experience pool, none of them an evaluation task)}: an LLM writes a candidate prompt, an LLM judge scores each extraction from 1 to 5 on coverage, specificity, discriminability, conciseness, and structural consistency, and weak dimensions are revised for up to five iterations. Porting to a new domain therefore needs example trajectories, not prompt engineering, and the discovered schema matches or exceeds the hand-written one (\S\ref{sec:results-store}; Appendix~\ref{app:discovery}).

\paragraph{Measurement.}
\label{sec:method-utilization}
To test whether agents engage with what they receive, an offline probe extracts \emph{identifiers} (file paths and CamelCase or snake\_case names of at least four characters, minus a stoplist) from the exact injected text and records which appear in the agent's own text or tool-call inputs, excluding tool results and discounting identifiers the paired no-memory run already surfaced. We report \emph{coverage}, the share of injected identifiers a run references; substring matching establishes presence rather than causal use, so we read it as mechanism evidence. Resolution contrasts are paired by task and tested with the two-sided exact McNemar test\rev{, pooled over replicates; a per-task Wilcoxon signed-rank test on resolve counts leads to the same conclusions (Appendix~\ref{app:summaries})}; utilization measures use Cohen's $d$ and the two-sided Mann-Whitney test. With 99 tasks per condition we read a single contrast as directional \del{unless it holds in every replicate}\rev{and report per-replicate counts for every multi-replicate condition (Table~\ref{tab:rep-summary})}; full temporal measures are in Appendix~\ref{app:utilization}.
\label{sec:method-stats}

\section{Experimental setup}
\label{sec:setup}

\paragraph{Benchmark and agent.} SWE-ContextBench Lite \citep{zhang2026swecontextbench} supplies 99 repository-level bug-fixing tasks, a pool of 300 solved trajectories, and a label marking the trajectory related to each task, which separates retrieval quality from the effect of the retrieved trajectory. A run is \emph{resolved} when the benchmark's held-out tests pass on its patch. Every condition runs the same agent on all 99 tasks, built on the Strands framework with six tools (\texttt{bash}, \texttt{read}, \texttt{write}, \texttt{edit}, \texttt{grep}, \texttt{glob}) in a per-task Docker container, capped at 100 turns and a wall-clock limit. Conversation history is never truncated, so injected-context effects are not confounded with history eviction. Claude Sonnet 4.5 is the agent unless stated; \S\ref{sec:results-who} varies it across five models.

\paragraph{Outcome measures.} \emph{Resolution} is intention-to-treat over all 99 tasks, so a run ending without an applicable patch counts as unresolved. \emph{Match@1} is the fraction of tasks whose top-ranked trajectory is the labeled one, comparable to the benchmark's reported top-1 rate \citep{zhang2026swecontextbench}. Treating the label as uniquely correct understates retrieval quality, so resolution is primary and match@1 secondary.

\paragraph{Conditions.} A \emph{condition} is one memory configuration; all share the agent and differ only in what enters context and when. \emph{No memory} supplies the problem statement alone. Start-of-run conditions inject once: \emph{Wrong summary} (an unrelated summary), \emph{Free summary}/\emph{Free trajectory} (the agent browses and selects), and \emph{Oracle summary}/\emph{Oracle trajectory} (the labeled trajectory as summary or raw transcript). Holding the labeled trajectory fixed, \emph{Oracle phased document} varies \del{only} its stored form\rev{, with the same fields as \sysname{}'s document plus cautionary guidance and test commands ($\sim$540 tokens)}. \sysname{} injects the retrieved document \rev{(hand-written schema; the discovered schema is compared on retrieval in \S\ref{sec:results-store})} and applies the recall policy of \S\ref{sec:method-recall}; its ablations inject the record eagerly, on struggle, or never, and evict or pin it. \sysname{}, no-memory, and the recall ablations were run three times on Sonnet 4.5 (pooled 297 task-runs, exact McNemar); other conditions were run once and read as directional. Tasks are treated as independent, which is imperfect when several share a repository. Per-condition accounting is in Appendix~\ref{app:summaries}\rev{; two tasks have no oracle run because their labeled trajectory is absent from the benchmark's released pool}.

\begin{figure}[t]
\centering
\includegraphics[width=0.78\textwidth]{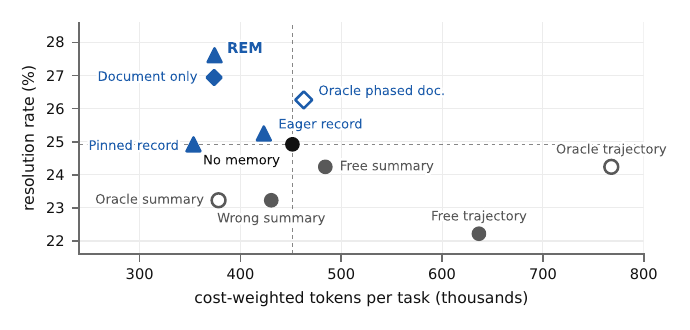}
\caption{\textbf{\del{Managing the lifecycle improves resolution and cost together; start-of-run injection improves neither.}\rev{Every configuration that injects the phased document matches or exceeds no-memory resolution, and the retrieved ones do so at lower cost; summaries and raw trajectories, even oracle ones, fall below it.}} Resolution against cost per task (Sonnet 4.5); up and to the left is preferable. Cost is weighted by token type (fresh $\times$1, cache write $\times$1.25, cache read $\times$0.1, output $\times$5\rev{, the agent model's relative list prices \citep{anthropic2026pricing}}). \del{Resolution is intention-to-treat over all 99 tasks.} \rev{Blue: phased document (diamond: document alone; triangle: document and implementation record). Gray: summary or raw trajectory. Hollow: oracle. Black, dashed: no memory. Filled blue points and no memory are three-replicate means; hollow and gray points are single runs.}}
\label{fig:main}
\end{figure}

\section{Results}
\label{sec:results}

\subsection{\sysname{} resolves more tasks at lower cost}
\label{sec:results-main}

\paragraph{Start-of-run \del{injection does}\rev{summaries and trajectories do} not move resolution.} The five \del{start-of-run}\rev{summary and trajectory} conditions resolve 22--24 of 99 tasks against 23 for no memory \rev{in the same replicate (24.9\% over three; Figure~\ref{fig:main})}, including the oracle conditions that receive the labeled trajectory, so relevance alone does not help. \rev{The benchmark's oracle-summary gain (34.3\% vs.\ 26.3\%; \citealp{zhang2026swecontextbench}) does not reproduce under our agent (23.2\% vs.\ 23.2\%).} \rev{The phased form is the exception: the same oracle trajectory rendered as a phased document resolves 26 in a single run (Figure~\ref{fig:main}; Table~\ref{tab:summary-sonnet-45}), consistent with the stored form, not relevance, driving the gain (\S\ref{sec:results-store}--\ref{sec:results-when}).} Cost still moves: a raw oracle transcript raises cost-weighted tokens per task from 452K to 768K and shortens runs from 70 to 50 turns, with unappliable patches up from 7 to 11, so the agent trades its own exploration for the transcript without resolving more.

\begin{table}[t]
\begin{center}
\small
\begin{tabular}{@{}llcc@{}}
\toprule
\textbf{Method} & \textbf{Stored unit} & \textbf{Candidates} & \textbf{Match@1} \\
\midrule
Free trajectory$^\dagger$ & raw trajectory ($\sim$\del{25{,}634}\rev{105K} tok) & self-selected & 22.2\% \\
Supermemory & managed chunks ($\sim$30 tok) & 15 & 23.2\% \\
Mem0 & extracted facts ($\sim$12 tok) & 3 & 25.3\% \\
Free summary$^\dagger$ & trajectory summary ($\sim$373 tok) & self-selected & 35.4\% \\
OpenViking & indexed file chunks ($\sim$246 tok) & 3 & 38.4\% \\
LangMem & extracted memory items ($\sim$48 tok) & 10 & 55.6\% \\
\midrule
\sysname{}, hand-written schema & phased document ($\sim$270 tok) & 5 & \textbf{66.7\%} \\
\sysname{}, discovered schema & phased document ($\sim$\del{270}\rev{305} tok) & 5 & \textbf{68.7\%} \\
\bottomrule
\end{tabular}
\end{center}
\caption{\textbf{The phase-structured document is the most accurate retrieval unit, ahead of every existing memory system.} Match@1 over the 99 tasks, after reranking for \sysname{}. Numbers for existing systems from \citet{zhang2026swecontextbench}\rev{, under each system's own retrieval configuration; our token counts are approximated as one token per four characters}. $^\dagger$From our runs, agent searches the pool.}
\label{tab:retrieval}
\end{table}

\paragraph{\sysname{} improves resolution and cost at once.} It resolves 27.6\% against 24.9\% without memory (+2.7 points, +11\% relative), with per-replicate counts of 27, 29, 26 exceeding no-memory's 23, 26, 25 in every replicate; against the strongest \del{start-of-run}\rev{summary or trajectory} condition the margin is 3.4 points. Of 297 paired task-runs, 20 resolve under \sysname{} that fail without memory and 12 the reverse. At 99 tasks this paired difference is not significant (exact McNemar $p = 0.22$), so we rest the resolution claim on its consistency across every replicate rather than on a single contrast, and treat the robust wins as the retrieval margin (\S\ref{sec:results-store}), which far exceeds the few-point run-to-run noise, and the cost reduction. Cost falls 17\%, from 452K to 374K tokens per task, with six fewer turns\rev{, mean wall-clock time down from 11.9 to 8.8 minutes,} and unappliable patches down from 7 to 4 (Appendix~\ref{app:outcomes}). \del{No other condition attains both the highest resolution and the lowest cost.}\rev{Every retrieved phased-document configuration matches or improves on no memory in both respects, and the document alone already resolves 26.9\% at essentially \sysname{}'s cost (Figure~\ref{fig:main}; \S\ref{sec:results-when}).} Correct content injected at the start \del{yields neither}\rev{as a summary or trajectory does not raise resolution}, whereas \del{deciding when it enters and leaves yields both}\rev{the phased document yields both, with or without the recall policy}: what converts experience into resolved tasks is \del{management of the lifecycle, not the relevance of what is stored}\rev{the form in which it is stored and the management of its lifecycle, not the relevance of what is retrieved}.

\subsection{\sysname{} retrieves the related trajectory more accurately than existing state-of-the-art systems}
\label{sec:results-store}

\begin{table}[t]
\begin{center}
\small
\begin{tabular}{@{}lrrrr@{}}
\toprule
\textbf{Injected form} & \textbf{Tokens} & \textbf{Injected} & \textbf{Referenced} & \textbf{Coverage} \\
\midrule
Wrong summary & $\sim$\del{217}\rev{380} & 8.9 & 0.3 & 3.4\% \\
Free trajectory & $\sim$\del{25{,}634}\rev{105K} & 67.7 & 17.9 & 26.7\% \\
Oracle trajectory & $\sim$\del{25{,}634}\rev{111K} & 67.7 & 18.1 & 27.2\% \\
Oracle summary & $\sim$\del{217}\rev{390} & 9.3 & 4.3 & 45.7\% \\
Free summary & $\sim$373 & 9.3 & 4.8 & 52.5\% \\
Oracle phased document & $\sim$\del{270}\rev{540} & 12.4 & 7.8 & 64.7\% \\
\sysname{} phased document & $\sim$270 & 11.0 & 7.5 & \textbf{69.1\%} \\
\midrule
Resolved vs.\ unresolved, $d$ ($p$) & & $+0.10$ (0.84) & $+0.18$ (0.10) & $+0.20^{*}$ (0.03) \\
\bottomrule
\end{tabular}
\end{center}
\caption{\textbf{The phased document is read more completely than any other stored form, and coverage tracks resolution.} \rev{Tokens are the stored unit's size, approximated as one token per four characters; the agent's file reader returned about half of each oracle trajectory ($\sim$54K).} Identifiers injected/referenced are per-run means; coverage is their ratio (Sonnet 4.5). The final row is the standardized mean difference between resolved and unresolved runs (two-sided Mann-Whitney $p$), pooled over the four start-of-run conditions ($n = 91$ resolved, 284 unresolved); $^{*}p < 0.05$. \sysname{} row is a mean over three replicates; others are one.}
\label{tab:utilization-main}
\end{table}

We score the phased document as a retrieval unit against the four state-of-the-art memory systems benchmarked on SWE-ContextBench Lite and the two agent self-search settings, by match@1 over the same 99 problem statements. Among these existing systems, match@1 tracks neither the stored-unit size nor the candidate budget (Table~\ref{tab:retrieval}): LangMem leads at 55.6\% from ten candidates, Supermemory reaches 23.2\% from fifteen, and self-search over raw trajectories is weakest at 22.2\%.

\sysname{} attains \textbf{68.7\%} match@1, 13.1 points above the strongest existing system and at least 30 above every other, a margin far larger than the few-point run-to-run variation at 99 tasks. Both stages contribute: cosine similarity selects the labeled trajectory for 62.6\% of tasks\rev{, already 7 points above LangMem before any rerank,} and the rerank raises this to 68.7\%, realizing 87\% of first-stage recall (Appendix~\ref{app:retrieval}). It also needs a smaller budget, recovering the labeled trajectory within five candidates for 78.8\% of tasks where LangMem needs ten for 73.3\%. Replacing the hand-written schema with the discovered one moves match@1 from 66.7\% to 68.7\%, so the write path needs no manual prompt engineering. The first-stage primitive, cosine similarity over text embeddings, is the same one Mem0 and LangMem use, which places the difference in the stored unit\rev{. The size-matched contrast points the same way: OpenViking returns chunks of $\sim$246 tokens, close to the phased document's 270, yet reaches 38.4\%, suggesting that structure rather than length is what retrieves}: factoids of 12--48 tokens discard connective structure and raw trajectories of $\sim$\del{25K}\rev{105K} tokens bury the identifying tokens, while a phased document of a few hundred tokens keeps what identifies a run and discards the search that produced it.

\subsection{Agents read the phased document more completely than any other stored form}
\label{sec:results-utilization}

\begin{table}[t]
\begin{center}
\small
\begin{tabular}{@{}lcccr@{}}
\toprule
\textbf{Recall policy for the record} & \textbf{Injected in} & \textbf{Resolution (sd)} & $\Delta$ ($p$) & \textbf{Cost tok/task} \\
\midrule
Never, document only & 0\% & 26.9\% (1.5) & $-0.7$ (0.82) & 374K \\
Eager, at first source edit & 100\% & 25.3\% (1.7) & $-2.3$ (0.17) & 423K \\
On struggle, then pinned & 56\% & 24.9\% (1.5) & $-2.7$ (0.08) & \textbf{353K} \\
\midrule
On struggle, then evicted (\sysname{}) & 58\% & \textbf{27.6\%} (1.5) & --- & 374K \\
\bottomrule
\end{tabular}
\end{center}
\caption{\textbf{Both gating on struggle and evicting after use improve resolution over the \del{alternatives}\rev{other record policies}.} Every row injects the same retrieved document and varies only the record policy (Sonnet 4.5). Resolution is the mean over three replicates of 99 tasks, sd across replicates in parentheses. $\Delta$ is the difference from \sysname{} in points, exact McNemar $p$ over 297 paired task-runs; none reaches $p<0.05$ at this sample size. Per-replicate counts in Table~\ref{tab:rep-summary}.}
\label{tab:ablation}
\end{table}

To explain \sysname{}'s advantage we test whether it comes from more complete use of the injected context, holding the labeled trajectory fixed and varying its stored form (Table~\ref{tab:utilization-main}). Uptake separates the forms sharply. The unrelated summary is nearly ignored at 3.4\%, confirming the probe tracks use rather than coincidence. The raw trajectory is referenced least (26.7--27.2\%), prose summaries reach 45.7--52.5\%, and the phased document reaches 64.7\% from an oracle and 69.1\% under \sysname{}. Two comparisons locate the cause: \del{at near-equal budget ($\sim$217 vs.\ 270 tokens) the phased document is read 1.5$\times$ as completely as the oracle summary}\rev{\sysname{}'s phased document is smaller than the oracle summary ($\sim$270 vs.\ 390 tokens) yet read 1.5$\times$ as completely (69.1\% vs.\ 45.7\%)}, so \emph{organization, not volume}, drives uptake; and the \del{raw}\rev{oracle} trajectory offers \del{six}\rev{5.5} times as many identifiers \rev{as the oracle phased document} at \del{ninety}\rev{about 200} times the tokens yet returns only \del{2.4}\rev{2.3} times as many referenced. Coverage also tracks success: among the four start-of-run forms, resolved runs reference more of what they were given than unresolved runs ($d = +0.20$, $p = 0.03$), an association carried by how much of the context a run takes up rather than how much it was given\rev{; under \sysname{}'s recall policies, which all inject the same phased document, coverage does not separate outcomes ($d = -0.06$, $p = 0.34$)} (Appendix~\ref{app:utilization}). Uptake is thus what the write path should maximize, and the phased document maximizes it.

\subsection{Gating on struggle and evicting after use each pay off}
\label{sec:results-when}

\begin{figure}[t]
\centering
\includegraphics[width=0.70\textwidth]{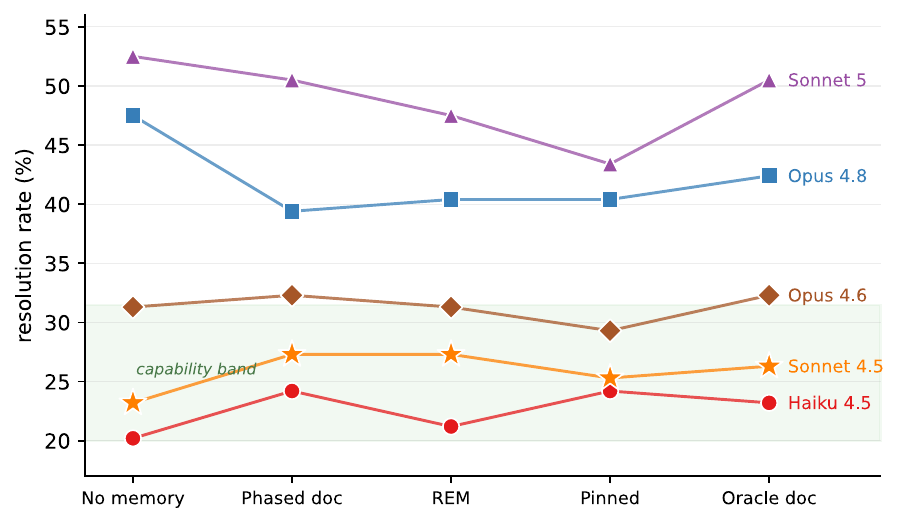}
\caption{\textbf{Memory helps agents in a capability band and costs the strongest ones.} Resolution by memory condition for five agent models, one replicate each\rev{ (the first for Sonnet 4.5)}. \emph{Phased doc} injects the retrieved document; \emph{REM} adds the struggle-gated, evicted record; \emph{Pinned} keeps that record; \emph{Oracle doc} substitutes the labeled trajectory's document. Shading marks no-memory rates at or below which \del{memory}\rev{the phased document} does not reduce resolution.}
\label{fig:sweep}
\end{figure}

Holding the retrieved document fixed, we vary only the policy for the implementation record (Table~\ref{tab:ablation}\rev{; Figure~\ref{fig:main}}). \emph{Gating beats eager injection}: eager resolves 25.3\% against 27.6\% for \sysname{}, a 2.3-point gap that never reverses across replicates\rev{ (the two tie in the third)}; the policies disagree on 19 of 297 task-runs and gating resolves 13 of them, while injecting the record in 58\% of runs rather than \del{essentially} all lowers cost from 423K to 374K tokens. \emph{Eviction beats pinning}, the most consistent contrast: pinning resolves 24.9\% against 27.6\% in all three replicates, and since the two policies draw near-identical coverage (50.3\% vs.\ 50.5\%), what eviction changes is not how much the agent reads but how long the record shapes later turns. The record's net contribution is nonetheless small: never injecting it resolves 26.9\%, within 0.7 points of \sysname{}\rev{ and tied with it in the first replicate}, so the gate and eviction mainly keep a costly second tier from doing harm, while the source of the advantage is the stored form (\S\ref{sec:results-store}--\ref{sec:results-utilization}). \rev{The model sweep agrees: the document alone matches or exceeds full \sysname{} on three of the four other models (Appendix~\ref{app:summaries}), so across agents the record rarely adds to what the document provides.}

\subsection{Memory helps agents in a capability band and costs the strongest ones}
\label{sec:results-who}

Repeating the comparison across five models spanning 20--53\% unaided resolution, with memory and policy fixed, the effect's sign depends on the recipient (Figure~\ref{fig:sweep}). The three models at or below 31.3\% never lose resolution from the phased document, gaining 4.0, 4.1, and 1.0 points\rev{ (Sonnet 4.5: first replicate, like the other single runs; 2.0 points over three)}, whereas the two above 47\% lose under every memory condition, by 2.0--9.1 points, with Opus 4.8 falling from 47.5\% to 39.4\%. Substituting the oracle document preserves the pattern, so the reversal is a property of the recipient, not a retrieval failure: content that orients a mid-capability agent displaces the reasoning of one that would have localized the defect unaided. Within the band, Haiku 4.5 gains 4.0 points from the document alone but only 1.0 with the record added, suggesting a record of another run's edits is usable only by an agent able to adapt rather than transcribe it. The sweep is one replicate per model, so the claim rests on the sign pattern; an experience memory is an intervention whose sign depends on the recipient, and whether to attach memory should condition on the agent as much as on the memory.

\section{Discussion and limitations}
\label{sec:discussion}

Coding agents produce far more experience than they reuse, and the field's response has been to build better stores and retrievers, on the assumption that reuse fails because the right run is hard to find. Our results point elsewhere: once an agent is running, what matters is the form its experience was written in, whether it needs that experience now, and how long it should stay in view. Two insights carry beyond \sysname{}. First, the stored form determines how much experience is used: the same run, at the same token cost, is read far more completely when organized by the phases of the work, and what separates a successful run from a failed one is how much of what it received it takes up. Second, more memory is not better memory: the gains from delaying and withdrawing costly context came from avoiding harm, and the same content that helped mid-capability agents cost the strongest ones several points.

Because \sysname{} treats memory as a set of swappable policy slots rather than a fixed store, its lessons are not tied to our implementation. The stored form is decoupled from the search algorithm, so a phased document can front existing retrievers; the recall and eviction decisions are model-agnostic and apply to any agent with an observable think-act loop; and because the extraction schema is discovered rather than hand-written, the pipeline extends to other diagnostic domains, such as text-to-SQL repair or incident response, from example trajectories rather than new engineering. Cast as a control policy over three decisions, an experience memory becomes something one can measure, price, and improve one decision at a time. These three decisions are exactly the operations a continually learning agent must eventually automate over its own growing experience---what to consolidate, when to surface it, and what to retire---so a mechanism that measures each in isolation is a prerequisite for memory that safely compounds.

\rev{In practice, \sysname{} cuts cost-weighted tokens by 17\%, mean wall-clock time from 11.9 to 8.8 minutes per task, and unappliable patches from 7.0 to 4.0 (Table~\ref{tab:outcomes}). Injecting a $\sim$270-token document rather than the raw run also shrinks what re-enters the agent's context, reducing the exposure of credentials or personal data captured in agent traces. And because the effect of memory depends on the recipient (\S\ref{sec:results-who}), a short no-memory pilot per agent model indicates when to attach it.}

\paragraph{\rev{Limitations.}}\sysname{} sets a new state of the art for experience reuse on SWE-ContextBench Lite, improving both retrieval and end-task resolution while lowering cost, and it does so with deliberately simple components, which leaves several promising directions open. \sysname{} performs \emph{single-pass} reuse from a fixed pool; letting an agent's own completed runs re-enter the store, with staleness and contradiction handled, is the natural path toward memory that compounds over time and, ultimately, toward continual learning for agents \citep{hu2026continual}. The struggle signal is a single hand-written observable, so a calibrated trigger\rev{ drawing on richer signals, such as repeated failed edits or lack of progress,} that spends a bounded budget of disturbance on runs that would have succeeded unaided is a clear next step. \rev{Eviction removes the record itself, but the agent's own response to it remains in the history, so a stricter variant could also prune that response.} The utilization probe establishes presence by substring match rather than causal use, inviting richer attribution. Most conditions beyond \sysname{} and its ablations were run once, so several secondary contrasts are directional and invite further replication. \rev{Retrieval numbers for existing systems are those reported by the benchmark under each system's own configuration, so a re-run through a shared embedder and reranker would further isolate the stored unit. Every task's labeled trajectory comes from the same repository as the task (99 of 99), and we did not measure how closely its patch resembles the gold fix, nor how stored fixes age as a codebase evolves.} And while we evaluate end-to-end on repository-level coding, the automatically discovered schema already ports to text-to-SQL (Appendix~\ref{app:discovery}), so validating the full pipeline in a second domain is an immediate opportunity.

\section{Conclusion}
\label{sec:conclusion}

We recast experience memory for long-running agents as a control problem rather than a lookup problem: a policy over three decisions---what to store, when to recall, and what to forget---each of which can be measured, priced, and replaced independently, and each of which we show is worth something on its own. \sysname{} realizes this view with a single phase-structured store, a struggle-gated recall, and one-shot eviction, and attains state-of-the-art retrieval and improved end-task resolution at lower cost on a public benchmark.

\begin{ack}
\rev{We thank Jiayuan Zhu for providing details on the SWE-ContextBench dataset and paper \citep{zhang2026swecontextbench}, and the anonymous CLEA reviewers and area chair for comments that improved this paper.}

\rev{\textbf{Author disclosure statement.} The authors declare no competing interests or conflict of interest.}

\rev{\textbf{Funding information.} All work was performed while the authors were employed at Amazon, on in-house infrastructure. The views expressed in this article are solely the authors' and may not necessarily reflect Amazon's viewpoint.}
\end{ack}

\bibliography{references}
\bibliographystyle{plainnat}

\clearpage
\appendix
\setcounter{topnumber}{2}
\setcounter{bottomnumber}{1}
\setcounter{totalnumber}{3}
\renewcommand{\topfraction}{0.7}
\renewcommand{\textfraction}{0.2}
\renewcommand{\floatpagefraction}{0.5}

\section{\rev{Algorithm}}
\label{app:algorithm}

\begin{algorithm}[H]
\caption{\textbf{\sysname{} Framework.} $\oplus$ concatenates context, $\emptyset$ is empty context; the three commented lines are the three lifecycle decisions.}
\label{alg:rem}
\begin{algorithmic}[1]
\Require trajectory pool $\mathcal{T}$, task $q$, struggle signal $S$, agent policy $\pi$, turn budget $N$
\Statex \textbf{Write path} \Comment{offline, once per trajectory}
\State $\mathcal{P} \gets \emptyset$
\For{$\tau \in \mathcal{T}$}
  \State $d \gets \textsc{Render}(\textsc{Extract}(\textsc{Flatten}(\tau)))$, \enspace $r \gets \textsc{Record}(\tau)$ \Comment{\emph{what to store}}
  \State $\mathcal{P} \gets \mathcal{P} \cup \{(d, \textsc{Embed}(d), r)\}$
\EndFor
\Statex \textbf{Read path} \Comment{once, at task start}
\State $\mathcal{C} \gets \textsc{TopK}_{(d,e,r) \in \mathcal{P}} \cos(\textsc{Embed}(q), e)$
\State $(d^\star, r^\star) \gets \textsc{Rerank}(q, \mathcal{C})$
\Statex \textbf{Recall node} \Comment{in-loop, every turn}
\State $h_0 \gets \emptyset$, \enspace $g \gets \textsc{false}$
\For{$t = 1$ \textbf{to} $N$}
  \If{$\lnot g$ \textbf{and} $S(o_{1:t-1})$}
    \State $\rho \gets r^\star$, \enspace $g \gets \textsc{true}$ \Comment{\emph{when to recall}}
  \Else
    \State $\rho \gets \emptyset$ \Comment{\emph{what to forget}: $r^\star$ is never re-injected}
  \EndIf
  \State $c_t \gets q \oplus d^\star \oplus h_{t-1} \oplus \rho$ \Comment{$d^\star$ persists, $\rho$ does not}
  \State $(a_t, o_t) \gets \pi(c_t)$, \enspace $h_t \gets h_{t-1} \oplus (a_t, o_t)$
  \State \textbf{if} $a_t = \textsc{halt}$ \textbf{then break}
\EndFor
\end{algorithmic}
\end{algorithm}

\section{Injection mechanics}
\label{app:delivery}

The retrieved phased document is \del{prepended to the task prompt and remains in the fixed prefix}\rev{placed in a file that the task prompt instructs the agent to read first; the agent reads it as its first action in every run, and it remains in the history thereafter}. The implementation record (raw edits and verifications from the related trajectory) is capped at 12{,}000 characters. The struggle trigger fires on the run's first \emph{genuinely failed} validation, a test or check whose result carries a failure, with edits to test files filtered so the agent's own test-writing does not count. In the default (evicting) policy the record appears once, in the turn after the trigger, and is absent afterward; the persistent variant pins it for the rest of the run, and the eager variant attaches it at the first source edit.

\section{End-task outcomes by condition}
\label{app:outcomes}

\begin{table}[h]
\begin{center}
\begin{tabular}{@{}lccrrrr@{}}
\toprule
\textbf{Condition} & \textbf{Resolved / 99} & \textbf{ITT} & \textbf{Cost tok/task} & \textbf{Turns} & \textbf{Min} & \textbf{PatchNA} \\
\midrule
No memory & 23, 26, 25 & 24.9\% & 452K & 70.4 & 11.9 & 7.0 \\
Wrong summary$^\dagger$ & 23 & 23.2\% & 431K & 72.0 & 14.4 & 7 \\
Free trajectory$^\dagger$ & 22 & 22.2\% & 637K & 66.3 & 13.1 & 11 \\
Free summary$^\dagger$ & 24 & 24.2\% & 484K & 68.6 & 13.8 & 10 \\
Oracle summary$^\dagger$ & 23 & 23.2\% & 378K & 67.4 & 13.7 & 6 \\
Oracle trajectory$^\dagger$ & 24 & 24.2\% & 768K & \textbf{49.9} & 9.5 & 11 \\
\midrule
\sysname{} & \textbf{27, 29, 26} & \textbf{27.6\%} & \textbf{374K} & 64.6 & \textbf{8.8} & \textbf{4.0} \\
\bottomrule
\end{tabular}
\end{center}
\caption{End-task outcomes (Sonnet 4.5). Resolved counts per replicate; ITT is their mean over all 99 tasks. Cost, turns, minutes, and PatchNA (patch does not apply) are means over replicates. $^\dagger$One replicate; other rows three.}
\label{tab:outcomes}
\end{table}

\section{Retrieval details}
\label{app:retrieval}

Pool documents and the query are embedded with Amazon Titan Text Embeddings v2, and the five nearest by cosine similarity go to one LLM rerank call asking for the single candidate most likely related by shared root cause, area of code, or fix pattern, constrained to a candidate index. Table~\ref{tab:retrieval-stages} separates the two stages for both schemas. First-stage recall bounds the rerank, and conversion reports the fraction realized. The discovered schema has lower first-stage recall yet higher match@1, because its documents rerank more reliably and none of its replies were unparseable. \rev{The discovered-schema pool holds 267 of the 300 documents, because 33 extractions failed on API rate limits; the 8 tasks whose labeled document is among them count as misses in its 68.7\%.}

\begin{table}[h]
\begin{center}
\begin{tabular}{@{}lccccc@{}}
\toprule
\textbf{Extraction schema} & \textbf{Cosine top-1} & \textbf{Gold in top-5} & \textbf{Match@1} & \textbf{Conversion} & \textbf{Parse fail.} \\
\midrule
Hand-written & 65.7\% & 83.8\% & 66.7\% & 79.6\% & 5 \\
Discovered & 62.6\% & 78.8\% & \textbf{68.7\%} & \textbf{87.2\%} & \textbf{0} \\
\bottomrule
\end{tabular}
\end{center}
\caption{Retrieval by stage on the 99 tasks. Cosine top-1 is first-stage accuracy, gold in top-5 is first-stage recall at the rerank budget, match@1 is post-rerank accuracy, conversion is match@1 over gold-in-top-5, and parse failures fell back to the cosine top-1.}
\label{tab:retrieval-stages}
\end{table}

\section{Full run summaries}
\label{app:summaries}

Tables~\ref{tab:summary-haiku-45}--\ref{tab:summary-sonnet-5} report per-condition accounting behind every resolution rate, one table per agent model: resolved counts, evaluated patches, intention-to-treat and precision rates, mean wall clock, assistant turns, cost-weighted tokens per instance (fresh $\times$1, cache write $\times$1.25, cache read $\times$0.1, output $\times$5), and the failure breakdown of runs without an evaluated patch. All main-text rates are the ITT column. \rev{Every oracle condition has no run for two of the 99 tasks (\texttt{scikit-learn-25365} and \texttt{-25763}), whose labeled trajectory (\texttt{scikit-learn-26323}) is absent from the benchmark's released pool of 300; over the 97 evaluable tasks oracle rates are 0.5--1.0 points higher (Sonnet 4.5: oracle summary 23.7\%, oracle trajectory 24.7\%, oracle phased document 26.8\%), which leaves every comparison in the text unchanged.}

\begin{table}[h]
\begin{center}\resizebox{\textwidth}{!}{%
\begin{tabular}{@{}lrrrrrrl@{}}
\toprule
condition & res/patch/N & ITT & prec. & min & turns & tok/inst & failures \\
\midrule
No memory & 20/88/99 & 20.2\% & 22.7\% & 9.5 & 71.1 & 444,070 & error 8, max\_turns 3 \\
Oracle phased document & 23/91/99 & 23.2\% & 25.3\% & 10.6 & 72.1 & 442,141 & error 6 \\
\sysname{}, document only & 24/93/99 & 24.2\% & 25.8\% & 10.2 & 68.3 & 381,828 & error 4, max\_turns 2 \\
\sysname{}, gate without eviction & 24/96/99 & 24.2\% & 25.0\% & 10.0 & 66.6 & 385,833 & error 2, max\_turns 1 \\
\sysname{} & 21/92/99 & 21.2\% & 22.8\% & 9.7 & 68.7 & 419,678 & error 3, max\_turns 4 \\
\bottomrule
\end{tabular}}
\end{center}
\caption{Full run summary, Haiku 4.5 (\texttt{claude-haiku-4-5}).}
\label{tab:summary-haiku-45}
\end{table}

\begin{table}[h]
\begin{center}\resizebox{\textwidth}{!}{%
\begin{tabular}{@{}lrrrrrrl@{}}
\toprule
condition & res/patch/N & ITT & prec. & min & turns & tok/inst & failures \\
\midrule
No memory & 23/97/99 & 23.2\% & 23.7\% & 14.5 & 71.5 & 452,536 & max\_turns 1, no\_patch 1 \\
Wrong summary & 23/95/99 & 23.2\% & 24.2\% & 14.4 & 72.0 & 430,526 & max\_turns 2, no\_patch 2 \\
Free trajectory & 22/94/99 & 22.2\% & 23.4\% & 13.1 & 66.3 & 636,529 & context\_overflow 3, max\_turns 2 \\
Free summary & 24/98/99 & 24.2\% & 24.5\% & 13.8 & 68.6 & 484,142 & max\_turns 1 \\
Oracle summary & 23/97/99 & 23.2\% & 23.7\% & 13.7 & 67.4 & 378,265 & none \\
Oracle trajectory & 24/90/99 & 24.2\% & 26.7\% & 9.5 & 49.9 & 767,810 & context\_overflow 3, error 1, max\_turns 3 \\
Oracle phased document & 26/96/99 & 26.3\% & 27.1\% & 10.1 & 74.5 & 462,790 & max\_turns 1 \\
\sysname{}, document only & 27/99/99 & 27.3\% & 27.8\% & 9.1 & 65.7 & 375,712 & none \\
\sysname{}, eager record & 23/98/99 & 23.2\% & 23.5\% & 7.9 & 60.1 & 426,743 & error 1 \\
\sysname{}, gate without eviction & 25/98/99 & 25.3\% & 25.3\% & 7.3 & 62.6 & 356,285 & error 1 \\
\sysname{} & 27/96/99 & 27.3\% & 27.3\% & 7.7 & 65.1 & 378,583 & error 1, max\_turns 1, no\_patch 1 \\
\bottomrule
\end{tabular}}
\end{center}
\caption{Full run summary, Sonnet 4.5 (\texttt{claude-sonnet-4-5}). First replicate; three-replicate means for \sysname{} and its recall ablations are in Tables~\ref{tab:outcomes} and~\ref{tab:ablation}.}
\label{tab:summary-sonnet-45}
\end{table}

\begin{table}[h]
\begin{center}\resizebox{\textwidth}{!}{%
\begin{tabular}{@{}lrrrrrrl@{}}
\toprule
condition & res/patch/N & ITT & prec. & min & turns & tok/inst & failures \\
\midrule
No memory & 31/99/99 & 31.3\% & 31.3\% & 14.6 & 56.7 & 279,947 & none \\
Oracle phased document & 32/94/99 & 32.3\% & 34.0\% & 15.2 & 56.6 & 296,278 & error 1, max\_turns 2 \\
\sysname{}, document only & 32/97/99 & 32.3\% & 33.0\% & 13.2 & 50.5 & 230,611 & max\_turns 2 \\
\sysname{}, gate without eviction & 29/98/99 & 29.3\% & 29.6\% & 14.5 & 51.9 & 248,034 & max\_turns 1 \\
\sysname{} & 31/98/99 & 31.3\% & 31.6\% & 13.8 & 49.9 & 242,595 & max\_turns 1 \\
\bottomrule
\end{tabular}}
\end{center}
\caption{Full run summary, Opus 4.6 (\texttt{claude-opus-4-6}).}
\label{tab:summary-opus-46}
\end{table}

\begin{table}[h]
\begin{center}\resizebox{\textwidth}{!}{%
\begin{tabular}{@{}lrrrrrrl@{}}
\toprule
condition & res/patch/N & ITT & prec. & min & turns & tok/inst & failures \\
\midrule
No memory & 47/95/99 & 47.5\% & 49.5\% & 17.0 & 24.5 & 129,623 & error 4 \\
Oracle phased document & 42/93/99 & 42.4\% & 45.2\% & 16.9 & 32.5 & 191,712 & error 1, timeout 3 \\
\sysname{}, document only & 39/98/99 & 39.4\% & 39.8\% & 17.0 & 25.6 & 140,603 & error 1 \\
\sysname{}, gate without eviction & 40/98/99 & 40.4\% & 40.8\% & 17.0 & 25.9 & 137,940 & max\_turns 1 \\
\sysname{} & 40/97/99 & 40.4\% & 41.2\% & 16.7 & 24.4 & 136,395 & error 2 \\
\bottomrule
\end{tabular}}
\end{center}
\caption{Full run summary, Opus 4.8 (\texttt{claude-opus-4-8}).}
\label{tab:summary-opus-48}
\end{table}

\begin{table}[h]
\begin{center}\resizebox{\textwidth}{!}{%
\begin{tabular}{@{}lrrrrrrl@{}}
\toprule
condition & res/patch/N & ITT & prec. & min & turns & tok/inst & failures \\
\midrule
No memory & 52/98/99 & 52.5\% & 53.1\% & 25.3 & 49.0 & 330,975 & no\_patch 1 \\
Oracle phased document & 50/94/99 & 50.5\% & 53.2\% & 28.0 & 54.1 & 364,212 & error 1, max\_turns 1, timeout 1 \\
\sysname{}, document only & 50/98/99 & 50.5\% & 51.0\% & 26.3 & 50.6 & 326,477 & max\_turns 1 \\
\sysname{}, gate without eviction & 43/98/99 & 43.4\% & 43.9\% & 24.2 & 46.7 & 301,082 & error 1 \\
\sysname{} & 47/98/99 & 47.5\% & 48.0\% & 25.3 & 47.3 & 303,257 & no\_patch 1 \\
\bottomrule
\end{tabular}}
\end{center}
\caption{Full run summary, Sonnet 5 (\texttt{claude-sonnet-5}).}
\label{tab:summary-sonnet-5}
\end{table}

Table~\ref{tab:rep-summary} reports the same accounting for the three replicates behind every multi-replicate number, with the mean the body cites. The no-memory condition ranges 23--26 resolved and \sysname{} 26--29, which is why single contrasts are read as directional. \rev{A per-task Wilcoxon signed-rank test on resolve counts (0--3 per task) leads to the same conclusions as the pooled McNemar test: \sysname{} against no memory $p = 0.21$ (McNemar $0.22$), document only $0.53$ ($0.82$), eager record $0.15$ ($0.17$), and gate without eviction $0.06$ ($0.08$).}

\begin{table}[h]
\begin{center}\resizebox{\textwidth}{!}{%
\begin{tabular}{@{}lrrrrrr@{}}
\toprule
condition & res/patch/N & ITT & PatchNA & min & turns & tok/inst \\
\midrule
No memory, replicate 1 & 23/97/99 & 23.2\% & 8 & 14.5 & 71.5 & 452,536 \\
\quad replicate 2 & 26/95/99 & 26.3\% & 6 & 10.5 & 69.0 & 442,914 \\
\quad replicate 3 & 25/98/99 & 25.3\% & 7 & 10.8 & 70.7 & 459,158 \\
\quad \textit{mean} & \textit{24.7/96.7/99} & \textit{24.9\%} & \textit{7.0} & \textit{11.9} & \textit{70.4} & \textit{451,536} \\
\midrule
\sysname{}, document only, replicate 1 & 27/99/99 & 27.3\% & 7 & 9.1 & 65.7 & 375,712 \\
\quad replicate 2 & 28/98/99 & 28.3\% & 3 & 10.0 & 65.6 & 373,133 \\
\quad replicate 3 & 25/98/99 & 25.3\% & 5 & 10.1 & 65.2 & 372,660 \\
\quad \textit{mean} & \textit{26.7/98.3/99} & \textit{26.9\%} & \textit{5.0} & \textit{9.7} & \textit{65.5} & \textit{373,835} \\
\midrule
\sysname{}, eager record, replicate 1 & 23/98/99 & 23.2\% & 7 & 7.9 & 60.1 & 426,743 \\
\quad replicate 2 & 26/99/99 & 26.3\% & 4 & 11.0 & 61.3 & 428,311 \\
\quad replicate 3 & 26/99/99 & 26.3\% & 6 & 9.2 & 59.7 & 414,308 \\
\quad \textit{mean} & \textit{25.0/98.7/99} & \textit{25.3\%} & \textit{5.7} & \textit{9.4} & \textit{60.4} & \textit{423,121} \\
\midrule
\sysname{}, gate without eviction, replicate 1 & 25/98/99 & 25.3\% & 4 & 7.3 & 62.6 & 356,285 \\
\quad replicate 2 & 26/99/99 & 26.3\% & 4 & 11.3 & 62.1 & 344,551 \\
\quad replicate 3 & 23/97/99 & 23.2\% & 7 & 9.7 & 63.0 & 359,318 \\
\quad \textit{mean} & \textit{24.7/98.0/99} & \textit{24.9\%} & \textit{5.0} & \textit{9.4} & \textit{62.6} & \textit{353,385} \\
\midrule
\sysname{}, replicate 1 & 27/96/99 & 27.3\% & 4 & 7.7 & 65.1 & 378,583 \\
\quad replicate 2 & 29/99/99 & 29.3\% & 4 & 9.9 & 66.6 & 387,842 \\
\quad replicate 3 & 26/97/99 & 26.3\% & 4 & 8.7 & 62.1 & 356,402 \\
\quad \textit{mean} & \textit{27.3/97.3/99} & \textit{27.6\%} & \textit{4.0} & \textit{8.8} & \textit{64.6} & \textit{374,276} \\
\bottomrule
\end{tabular}}
\end{center}
\caption{Per-replicate accounting for the three-replicate conditions (Sonnet 4.5). Every replicate runs all 99 tasks. Columns: resolved/patches/instances, ITT rate, count of runs whose patch does not apply, and mean minutes, turns, and cost-weighted tokens per instance. Mean rows are the values the body reports.}
\label{tab:rep-summary}
\end{table}

\section{Utilization by condition}
\label{app:utilization}

Table~\ref{tab:utilization} reports the full utilization probe per condition (Sonnet 4.5), adding the temporal measures Table~\ref{tab:utilization-main} omits. Coverage denominators scale with injected size, so comparisons are cleanest within groups receiving the same amount, which is why \sysname{}'s policies are separated from the start-of-run conditions.

\begin{table}[h]
\begin{center}
\begin{tabular}{@{}lrrrrr@{}}
\toprule
\textbf{Condition} & $n$ & \textbf{Coverage} & \textbf{Window} & \textbf{Dead weight} & \textbf{First infl.} \\
\midrule
\sysname{}, phased document only & $3\times$99 & 69.1\% & 38.8 & 35.3\% & 4.1 \\
\sysname{}, gate and evict & $3\times$99 & 50.3\% & 42.9 & 27.2\% & 4.8 \\
\sysname{}, gate without eviction & $3\times$99 & 50.5\% & 42.1 & 28.1\% & 4.3 \\
\sysname{}, eager record & $3\times$99 & 38.8\% & 44.9 & 19.1\% & 3.8 \\
\midrule
Oracle phased document & 97 & 64.7\% & 45.6 & 35.7\% & 2.1 \\
Free summary & 97 & 52.5\% & 28.2 & 52.8\% & 3.6 \\
Oracle summary & 97 & 45.7\% & 27.4 & 51.9\% & 5.1 \\
Oracle trajectory & 97 & 27.2\% & 36.6 & 19.3\% & 4.0 \\
Free trajectory & 97 & 26.7\% & 44.5 & 23.0\% & 5.0 \\
Wrong summary & 99 & 3.4\% & 0.5 & 96.3\% & 8.7 \\
\bottomrule
\end{tabular}
\end{center}
\caption{Identifier utilization per condition (Sonnet 4.5). Window is the span in turns from first to last reference, dead weight the fraction of the run after the last reference, and first influence the turn of the first reference. \sysname{} rows are means over three replicates; start-of-run rows are one.}
\label{tab:utilization}
\end{table}

Table~\ref{tab:utilization-assoc} tests coverage against resolution and, because a share can move either way, its two component quantities separately. Among the start-of-run forms the association is carried by how many identifiers were referenced; under \sysname{}'s recall policies neither quantity separates the outcomes.

\begin{table}[h]
\begin{center}
\begin{tabular}{@{}llrrrr@{}}
\toprule
\textbf{Grouping} & \textbf{Quantity} & \textbf{Resolved} & \textbf{Unresolved} & $d$ & \textbf{MW $p$} \\
\midrule
\multirow{3}{*}{Start-of-run forms}
 & coverage & 0.423 & 0.378 & $+0.20$ & \textbf{0.03} \\
 & identifiers referenced & 12.9 & 11.0 & $+0.18$ & 0.10 \\
 & identifiers injected & 39.7 & 36.2 & $+0.10$ & 0.84 \\
\midrule
\multirow{3}{*}{\sysname{} recall policies}
 & coverage & 0.511 & 0.527 & $-0.06$ & 0.34 \\
 & identifiers referenced & 12.4 & 12.7 & $-0.02$ & 0.42 \\
 & identifiers injected & 28.8 & 28.3 & $+0.02$ & 0.76 \\
\bottomrule
\end{tabular}
\end{center}
\caption{Coverage against resolution, with its two component quantities tested separately. Coverage is identifiers referenced over injected. Start-of-run forms pool the oracle and free summary and trajectory conditions ($n = 91$ resolved, 284 unresolved); \sysname{} recall policies pool the four phased conditions over three replicates ($n = 311$ resolved, 868 unresolved). $d$ is the standardized mean difference; MW $p$ is two-sided Mann-Whitney. \rev{Runs without an evaluation verdict are excluded (13 start-of-run, 9 \sysname{}); start-of-run forms also exclude the two tasks without a labeled trajectory in the pool.}}
\label{tab:utilization-assoc}
\end{table}

\section{Schema discovery details}
\label{app:discovery}

The meta-loop runs four steps per iteration: an LLM writes or revises a structuring prompt (six to ten named fields with types, rules requiring exact names, and a JSON template), the candidate runs over a validation set, an LLM judge scores each extraction from 1 to 5 on coverage, specificity, discriminability, conciseness, and structural consistency, and the loop stops when every dimension reaches a mean of 4.0 or after five iterations. Bootstrap, revision, and judging use Claude Sonnet 4, and trajectories are truncated to 15K characters for the meta-prompt.

For repository-level coding the loop converged on its first iteration, with means over eight extractions of 4.12 (coverage), 4.00 (specificity), 4.75 (discriminability), 4.38 (conciseness), and 4.50 (structural consistency); the prompt is $\sim$1.7K characters and defines eight fields. Discriminability, scored highest, matters most for a store whose documents must be told apart at retrieval time, and the discovered schema has the higher match@1 and better rerank conversion (Appendix~\ref{app:retrieval}). Table~\ref{tab:schema-fields} sets the discovered fields against the hand-written ones: three coincide, the discovered schema consolidates the code-structure and test fields, drops files merely read and the catch-all identifier list, and adds three fields describing the problem, spending fewer fields on where the fix landed and more on what the failure looked like.

\begin{table}[h]
\begin{center}
\resizebox{\textwidth}{!}{%
\begin{tabular}{@{}ll@{}}
\toprule
\textbf{Hand-written} & \textbf{Discovered} \\
\midrule
\texttt{files\_modified} & \texttt{files\_modified} \\
\texttt{root\_cause} & \texttt{root\_cause} \\
\texttt{fix\_approach} & \texttt{fix\_approach} \\
\texttt{functions\_changed}, \texttt{classes\_involved} & \texttt{related\_functions} \\
\texttt{test\_names}, \texttt{test\_commands} & \texttt{test\_cases} \\
\texttt{files\_read}, \texttt{key\_identifiers} & --- \\
--- & \texttt{bug\_description}, \texttt{repository\_info}, \texttt{error\_messages} \\
\bottomrule
\end{tabular}}
\end{center}
\caption{Extraction fields, hand-written against discovered, for repository-level coding. Rows pair fields serving the same purpose; the last two list fields present in only one schema.}
\label{tab:schema-fields}
\end{table}

Running the same loop unchanged on text-to-SQL trajectories converged on the first iteration (means of 5.0 for coverage, specificity, discriminability, and structural consistency, 4.8 for conciseness) and produced a ten-field schema covering the database, schemas and tables, columns, SQL errors and data-quality problems, analytical approach, answer, query patterns, and business context. None of the coding fields survive, which is intended, since the pipeline's only domain-specific artifact is the schema.

\end{document}